\documentclass{article}

\usepackage{PRIMEarxiv}

\usepackage[utf8]{inputenc} 
\usepackage[T1]{fontenc}    
\usepackage{hyperref}       
\usepackage{url}            
\usepackage{booktabs}       
\usepackage{amsfonts}       
\usepackage{nicefrac}       
\usepackage{microtype}      
\usepackage{lipsum}
\usepackage{fancyhdr}       
\usepackage{graphicx}       
\graphicspath{{media/}}     

\title{Domain Specific Wav2vec 2.0 Fine-tuning For The SE\&R 2022 Challenge}

\author{
  Alef Iury Siqueira Ferreira \\
  Institute of Informatics \\
  Federal University of Goiás \\
  Brazil\\
  \texttt{alef\_iury\_c.c@discente.ufg.br} \\
   \And
  Gustavo dos Reis Oliveira \\
  Institute of Informatics \\
  Federal University of Goiás \\
  Brazil\\
  \texttt{gustavo\.reis2\_c.c@discente.ufg.br} \\
}

\begin{document}
\maketitle

\begin{abstract}
This paper presents our efforts to build a robust ASR model for the shared task \textit{Automatic Speech Recognition for spontaneous and prepared speech \& Speech Emotion Recognition in Portuguese (SE\&R 2022)}. The goal of the challenge is to advance the ASR research for the Portuguese language, considering prepared and spontaneous speech in different dialects. Our method consist on fine-tuning an ASR model in a domain-specific approach, applying gain normalization and selective noise insertion. The proposed method improved over the strong baseline provided on the test set in 3 of the 4 tracks available.
\end{abstract}



\section{Introduction}

The performance of Automatic Speech Recognition systems (ASRs) has increased significantly with the development of modern neural network topologies and the use of massive amount of data to train the models \cite{li2021recent}. Although the accuracy of recent models improved for high-resource languages, such as English, the development of ASR models in other languages is still a difficult task using the same technologies \cite{amodei2016deep, quintanilha2020open}. In this scenario, Self-Supervised Learning (SSL), a method in which representations with semantic information are learned by using unlabelled data, emerged as an important advance, allowing the training of deeper models using less labelled data \cite{baevski2019vq, jaiswal2021survey}. In this line of work, this paper explores the use of the Wav2vec 2.0 \cite{baevski2020Wav2vec}, a framework for self-supervised learning of discrete representations from raw audio data.

\begin{figure}[ht]
	\centering
	\includegraphics[width=0.6\columnwidth]{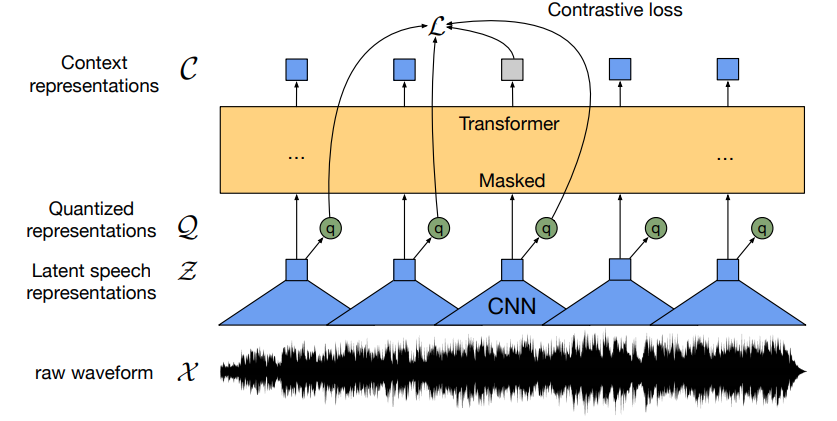}
	\caption[Wav2vec 2.0]{Illustration of the Wav2vec 2.0 framework \cite{baevski2020Wav2vec}.}
\label{fig:arquitetura-Wav2vec2}
\end{figure}

Wav2vec 2.0 (Figure \ref{fig:arquitetura-Wav2vec2}) is inspired by previous works in unsupervised pre-training for speech recognition, that is, Wav2vec \cite{schneider2019Wav2vec} and Vq-Wav2vec \cite{baevski2019vq}. During pre-training the model learns speech representations solving a contrastive task which requires identifying the correct quantized latent speech representations of a masked time step among a set of distractors. After the self-supervised pre-training, the model can be fine-tuned on labeled data in a supervised task, like ASR, adding a randomly initialized linear projection on top of the context network with $N$ classes and a loss function specific to the task at hand, like CTC, for instance.

The model shows important results among low resource languages. In Portuguese, for example, \cite{gris2021brazilian} and \cite{c2021coraa} demonstrated that the fine-tuning of the Wav2vec 2.0 model achieves state-of-the-art (SOTA) results only using publicly available datasets.

An important aspect to consider when training an ASR model is the quality and the domain of the data \cite{ seltzer2013investigation, likhomanenko2020rethinking, hsu2021robust}. While most of the available public datasets are composed of prepared speech \cite{c2021coraa}, mostly read sentences \cite{alencar2008lsf, Pratap_2020}, the domain of real ASRs are far more complex, mainly because it is formed by spontaneous speech and different speech dialects. Quality is another issue: most of ASR use cases involve high noise environments or low recording equipment, which is not adressed in most of the public datasets available \cite{c2021coraa}.

To stimulate research that can advance the present SOTA in ASR in Portuguese, for both prepared and spontaneous speech, the shared-task \textit{Automatic Speech Recognition for spontaneous and prepared speech \& Speech Emotion Recognition in Portuguese (SE\&R 2022)} introduces a new baseline for ASR and a new dataset in Portuguese \cite{c2021coraa}. The Corpus of Annotated Audios (CORAA ASR), a large corpus of spontaneous and prepared speech, is composed by various subsets in Portuguese with different characteristics. The baseline achieves a Word Error Rate (WER) of 24.18\% on CORAA ASR test set, a difficult dataset containing samples with low quality, noise, and a variety of domains and dialects.

In this work, we investigate the fine-tuning of the baseline model \cite{c2021coraa} proposed by the shared task, a fine-tuned model based on the Wav2vec 2.0 XLSR-53 \cite{conneau2020unsupervised}, using only public available Portuguese datasets, including the CORAA ASR dataset. We conducted several experiments in different domains for the challenge and explored the use of selective noise insertion and audio normalization during training. This work is organized as follows: Section \ref{sec:methods} discuss the proposed methods and Section \ref{sec:results}
shows and discusses the obtained results. Finally, Section \ref{sec:conclusions} presents the conclusions of this work.

\section{Methods}\label{sec:methods}

\subsection{Datasets}

We used several publicly available datasets in Portuguese. Besides CORAA ASR, most of them is composed by prepared speech. In general, we opted to use all the data in the gathered datasets for training, except the dev part of the CORAA ASR, as presented in Table \ref{tab:split}. The datasets used in this work are:

\begin{itemize}
    \item CETUC \cite{alencar2008lsf}: contains approximately 145 hours of Brazilian Portuguese speech distributed among 50 male and 50 female speakers, each pronouncing approximately 1,000 phonetically balanced sentences selected from the CETEN-Folha\footnote{\url{https://www.linguateca.pt/cetenfolha/}} corpus;
    
    \item Common Voice (CV) 7.0 \cite{ardila2020common}: is a project proposed by Mozilla Foundation with the goal to create a wide open dataset in different languages. In this project, volunteers donate and validate speech using the official site\footnote{\url{https://commonvoice.mozilla.org/pt}};
    
    \item Multilingual LibriSpeech (MLS) \cite{Pratap_2020}: a massive dataset available in many languages. The MLS is based on audiobook recordings in public domain like LibriVox\footnote{\url{https://librivox.org/}}. The dataset contains a total of 6k hours of transcribed data in many languages. The set in Portuguese used in this work\footnote{\url{http://www.openslr.org/94/}} (mostly Brazilian variant) has approximately 284 hours of speech, obtained from 55 audiobooks read by 62 speakers;
    
    \item Multilingual TEDx \cite{salesky2021multilingual}: a collection of audio recordings from TEDx talks in 8 source languages. The Portuguese set (mostly Brazilian Portuguese variant) contains 164 hours of transcribed speech. 
    
    \item Corpus of Annotated Audios (CORAA ASR) v1 \cite{c2021coraa}: is a public available dataset that contains 290.77 hours of validated pairs (audio-transcription) in Portuguese (mostly Brazilian Portuguese variant) and is comprised by five other corpora: ALIP \cite{Goncalves_2019}, C-ORAL Brasil I \cite{c_oral}, NURC-Recife \cite{NURC}, SP2010 \cite{sp2010} and TEDx Portuguese talks.
    
\end{itemize}

\begin{table*}[!h]
\centering \scriptsize
\begin{tabular}{l|r|r|r|r|r}    
\toprule
\textbf{Dataset} &  \textbf{Subset}                                 &  \textbf{Type}& \textbf{Train}  & \textbf{Dev (validation)}     & \textbf{Test} \\
\midrule
Others & CETUC                              & Prepared Speech & 144.65h            & --                  & --           \\
                                    & Common Voice                      & Prepared Speech & 112.08h         & --                 & --           \\
                                    & MLS (Portuguese)                                & Prepared Speech & 168.34h        & --                    & --          \\
                                    & Multilingual TEDx (Portuguese)     & Prepared Speech & 152.17h        & --                    & --          \\
\midrule
CORAA ASR\cite{c2021coraa}             & ALIP                          & Spontaneous Speech & 33.40h            & 0.99h               &   1.57h           \\
                                    & C-ORAL Brasil I               & Spontaneous Speech & 6.54h            & 1.13h               &   1.97h           \\
                                    & NURC-Recife                   & Spontaneous Speech & 137.08h          & 1.29h               &   2.94h           \\
                                    & SP2010                        & Spontaneous Speech & 27.83h           & 1.13h               &   2.18h           \\
                                    & TEDx Portuguese               & Prepared Speech & 68.67h              & 1.37h               &   2.70h           \\
\midrule
\multicolumn{3}{l|}{\textbf{Total}}                                                 & 850.76h          & 5.91h     & 11.36h\\
\bottomrule   
\end{tabular} 
\caption{Dataset splits used in this work.}
\label{tab:split}
\end{table*}

\subsection{Experiments}

Our experiments consists on the fine-tuning of the baseline model of \cite{c2021coraa}. For each experiment, we trained the model for 5 epochs, using a batch size of 192, and using Adam \cite{DBLP:journals/corr/KingmaB14} where the learning rate is $3^{e-05}$ that is warmed up for the first 400 updates, then linearly decayed for the remained. For the experiments, we used a NVIDIA TESLA V100 32GB, a NVIDIA TESLA Tesla P100 16GB and a NVIDIA A100 80GB, depending on the type of audio pre-processing used. The code to replicate the results is available at \url{https://github.com/alefiury/SE-R_2022_Challenge_Wav2vec2}. 

In total, we conducted five main experiments to test our methods:

\begin{itemize}

    \item Experiment 1: Wav2vec 2.0 XLSR-53 - Base: For this experiment, the model was fine-tuned considering the whole train set, but did not receive neither normalization nor noise addition;
    
    \item Experiment 2: Wav2vec 2.0 XLSR-53 - Norm: The model was fine-tuned with the whole train set with normalization. For the normalization, the mean gain of all the audios in the train set was considered;
    
    \item Experiment 3: Wav2vec 2.0 XLSR-53 - Norm and SNA: The model was fine-tuned with gain normalization and selective noise addition. The audios were normalized considering the mean gain of all the audios in the train set, and those audios pertaining to datasets that were considered to have a low presence of noise, namely MLS and CETUC, received randomly one of the following 5 possible types of noises: additive noise, being music or nonspeech noises from the MUSAN Corpus \cite{snyder2015musan}, Room impulse responses \cite{7953152}, Addition or reduction of gain, Pitch shift and Gaussian noise;
    
    \item Experiment 4: Wav2vec 2.0 XLSR-53 - Norm + Prepared Speech: Model fine-tuned based on the final trained model of Experiment 2, but considering just the prepared speech data from the CORAA ASR dataset, trained for more 5 epochs;
    
    \item Experiment 5: Wav2vec 2.0 XLSR-53 - Norm + Spontaneous Speech: Model fine-tuned based on the final trained model of Experiment 2, but considering just the spontaneous speech data from the CORAA ASR dataset, trained for more 5 epochs.
    
\end{itemize}

\section{Results and Discussion}\label{sec:results}

The shared-task consists of 4 tracks. Each track has a domain specific scenario, that includes prepared speech and spontaneous speech. In this regard, we conducted a prior analysis (Section \ref{subsec:dev}) using the dev set to select the best approaches based on 3 of the 4 tracks available: Mixed, Prepared Speech PT\_BR and Spontaneous Speech. The best models were selected and then submitted for evaluation. Our final results are presented in Section \ref{sec:final}. 

\subsection{Dev Set Analysis\label{subsec:dev}}

Overall, our models did not show a huge improvement in performance when compared to the baseline model. Even though we fine-tuned a model that is considered the state-of-the-art in Brazilian Portuguese, we suspect that the number of training epochs might have been insufficient to obtain an increase in performance, or that the baseline model might have already reached a local optima.

Furthermore, as presented in Table \ref{tab:devsubset}, the model that was fine-tuned using prepared speech clearly improved the results in the Prepared Speech subset (and consequently the Mixed subset). The same phenomenon could not be seen in the Spontaneous Speech subset. A possible explanation to this fact is that most of the data of the datasets that were added to the train set are comprised of prepared speech, which might have contributed to the increase in performance in this particular domain. Another possible explanation is the low number of training epochs used to train the models.

\begin{table*}[hbt!]
\centering \scriptsize
\begin{tabular}{l|c|c|c|c|c|c|c|c}    
\toprule
& \multicolumn{2}{|c|}{Prepared Speech PT\_BR} & \multicolumn{2}{|c|}{Prepared Speech PT\_PT} & \multicolumn{2}{|c|}{Spontaneous Speech} & \multicolumn{2}{|c}{Mixed} \\
\midrule
\textbf{Model} & \textbf{CER} &  \textbf{WER} & \textbf{CER} &  \textbf{WER} & \textbf{CER} &  \textbf{WER} & \textbf{CER} &  \textbf{WER}  \\
\midrule
Wav2vec 2.0 XLSR-53 Baseline & 4.53\% & 13.09\% & 17.73\% & 42.09\% & \textbf{14.60} \% & 30.13\% & 12.87\% & 28.86\%\\
\midrule
Wav2vec 2.0 XLSR-53 - Base & 5.65\% & 15.38\% & 16.82\% & 38.66\% & 15.13\% & 30.86\% & 13.18\% & 28.94\%\\
\midrule
Wav2vec 2.0 XLSR-53 - Norm & 5.43\% & 14.85\% & 17.07\% & 38.06\% & 14.77\% & \textbf{30.1\%} & 13.01\% & 28.28\%\\
\midrule
Wav2vec 2.0 XLSR-53 - Norm and SNA & 5.26\% & 14.46\% & 17.38\% & 38.80\% & 14.77\% & 30.12\% & 13.04\% & 28.37\%\\
\midrule
Wav2vec 2.0 XLSR-53 - Norm + Prepared Speech & \textbf{4.38\%} & \textbf{12.46\%} & \textbf{15.63\%} & \textbf{35.27\%} & 15.00\% & 30.83\% & \textbf{12.50\%} & \textbf{27.35\%}\\
\midrule
Wav2vec 2.0 XLSR-53 - Norm + Spontaneous Speech & 4.71\% & 13.38\% & 16.30\% & 37.44\% & 14.93\% & 30.36\% & 12.72\% & 27.89\%\\
\bottomrule   
\end{tabular} 
\caption{Dev set analisys by subset}
\label{tab:devsubset}
\end{table*}

Additionally, the noise insertion did not gave further improvement in performance. Nevertheless, the results of the SNA model in some more noisy subsets of the CORAA ASR dataset, like ALIP and NURC, for instance, showed some interesting and promising results when compared to the baseline. These results are shown in Table \ref{tab:devdatcoraa}.

\begin{table*}[hbt!]
\centering \scriptsize
\begin{tabular}{l|c|c|c|c|c|c|c|c}    
\toprule
& \multicolumn{2}{|c|}{ALIP} & \multicolumn{2}{|c|}{NURC-Recife} & \multicolumn{2}{|c|}{C-ORAL-BRASIL I} &  \multicolumn{2}{|c}{SP2010}\\
\midrule
\textbf{Model} & \textbf{CER} &  \textbf{WER} & \textbf{CER} &  \textbf{WER} & \textbf{CER} &  \textbf{WER} & \textbf{CER} &  \textbf{WER}  \\
\midrule
Wav2vec 2.0 XLSR-53 Baseline & 15.08\% & 30.13\% & 14.02\% & 32.50\% & 20.09\% & 38.41\% & \textbf{9.21\%} & 19.49\% \\
\midrule
Wav2vec 2.0 XLSR-53 - Base & 15.42\% & 30.23\% & 14.27\% & 33.32\% & 21.23\% & 40.19\% & 9.58\% & 19.72\% \\
\midrule
Wav2vec 2.0 XLSR-53 - Norm & 14.98\% & 29.68\% & 14.23\% & 32.42\% & 20.47\% & 38.98\% & 9.38\% & 19.36\% \\
\midrule
Wav2vec 2.0 XLSR-53 - Norm and SNA & \textbf{14.97\%} & \textbf{29.24\%} & \textbf{14.00\%} & \textbf{32.26\%} & 20.68\% & 39.65\% & 9.42\% & \textbf{19.31\%} \\
\midrule
Wav2vec 2.0 XLSR-53 - Norm + Prepared Speech & 15.19\% & 30.36\% & 14.61\% & 33.78\% & 20.30\% & 39.30\% & 9.89\% & 19.88\% \\
\midrule
Wav2vec 2.0 XLSR-53 - Norm + Spontaneous Speech & 15.31\% & 29.84\% & 14.23\% & 32.67\% & \textbf{19.89\%} & \textbf{37.93\%} & 10.32\% & 21.01\% \\
\bottomrule   
\end{tabular} 
\caption{Dev set analisys by dataset in CORAA ASR}
\label{tab:devdatcoraa}
\end{table*}

\subsection{Final Results\label{sec:final}}
Table \ref{tab:testsubset} compares the baseline with our selected models in the test set. The model Wav2vec 2.0 XLSR-53 - Norm + Prepared Speech surpassed the strong baseline in the Prepared Speech PT\_BR, Prepared Speech PT\_PT and the Mixed tracks. As seen in the results based on the dev set, the fact that most of the data of the datasets that were added to the train set are comprised of prepared speech, might have contributed to the increase in performance in this domain in both Portuguese variants. Lastly, even though we were not able to surpass the baseline model in the Spontaneous Speech track, we achieved competitive results with both submitted models.

\begin{table*}[hbt!]
\centering \scriptsize
\begin{tabular}{l|c|c|c|c|c|c|c|c}    
\toprule
& \multicolumn{2}{|c|}{Prepared Speech PT\_BR} & \multicolumn{2}{|c|}{Prepared Speech PT\_PT} & \multicolumn{2}{|c|}{Spontaneous Speech} & \multicolumn{2}{|c}{Mixed} \\
\midrule
\textbf{Model} & \textbf{CER} &  \textbf{WER} & \textbf{CER} &  \textbf{WER} & \textbf{CER} & \textbf{WER} & \textbf{CER} &  \textbf{WER}  \\
\midrule
Wav2vec 2.0 XLSR-53 Baseline & 3.56\% & \textbf{11.19\%} & 17.08\% & 39.75\% & \textbf{12.39}\% & \textbf{26.24\%} & 11.35\% & 25.85\%\\
\midrule
Wav2vec 2.0 XLSR-53 - Norm & 4.40\% & 12.71\% & 15.33\% & 35.31\% & 12.51\% & 26.50\% & 11.18\% & 25.25\%\\
\midrule
Wav2vec 2.0 XLSR-53 - Norm + Prepared Speech & \textbf{3.55} \% & 11.25\%  & \textbf{15.16\%} & \textbf{34.68\%} & 12.61\% & 26.81\% & \textbf{10.98\%} & \textbf{24.89\%}\\
\bottomrule
\end{tabular} 
\caption{Test set analisys by subset}
\label{tab:testsubset}
\end{table*}

\subsection{Additional Experiments}
\label{sec:additionalResults}

After selecting and submitting the best results, we performed some additional experiments to further explore our proposed methods using the dev set. We tried to use text correction in the outputs of the ASR models, and train the normalized models for a longer period of time using early stopping considering the prepared speech and spontaneous speech data from the CORAA ASR dataset. 

The text correction was done with an additional post-processing step, using a KenLM \cite{heafield-2011-kenlm} model. For the different tasks, we used 2 different KenLM models: one for spontaneous speech, which was built using subsets of the CORAA ASR dataset containing spontaneous speech phrases. And the other one was built considering wikipedia in portuguese texts, as proposed by \cite{quintanilha2020open}. Both were 4-grams. We found that this post-processing of the predictions from the ASR models did not improve the results from the dev set on the Prepared Speech PT\_BR and Spontaneous Speech tracks, as can be seen in Table \ref{tab:adddevset}, in fact they were worse. One possible explanation is that some of the decoder hyper-parameters did not work well with our ASR models. Another possibility is that the 4-gram trained with spontaneous text was built with a small amount of text, which might have decreased the performance of the model. However, the results on the Prepared Speech PT\_PT track were much better compared to the previous experiments. This result suggests that the LM might improve results when there are few domain data used to train the Wav2vec model, since most of our training data was composed by Brazilian Portuguese audios. 

Furthermore, as we had suspected earlier, the model with gain normalization that were trained considering the spontaneous speech data from the CORAA ASR corpus and for a longer period of time performed better in its respective subtrack, strengthening our hypothesis that our results did not improve in the main experiments due to a low number of training epochs.

\begin{table*}[!h]
\centering \scriptsize
\begin{tabular}{l|c|c|c|c|c|c|c|c}    
\toprule
& \multicolumn{2}{|c|}{Prepared Speech PT\_BR} & \multicolumn{2}{|c|}{Prepared Speech PT\_PT} & \multicolumn{2}{|c|}{Spontaneous Speech} & \multicolumn{2}{|c}{Mixed} \\
\midrule
\textbf{Model} & \textbf{CER} &  \textbf{WER} & \textbf{CER} &  \textbf{WER} & \textbf{CER} &  \textbf{WER} & \textbf{CER} &  \textbf{WER}  \\
\midrule
Norm + Prepared Speech + Prepared KenLM & 5.59\% & 14.11\% & \textbf{15.23\%} & \textbf{32.36\%} & 16.26\% & 30.66\% & 13.33\% & 26.95\%\\
\midrule
Norm + Prepared Speech + Spontaneous KenLM & 5.83\% & 15.58\% & 16.52\% & 36.74\% & 16.73\% & 33.78\% & 13.95\% & 29.97\%\\
\midrule
Norm + Spontaneous Speech + Prepared KenLM & 5.80\% & 13.94\% & 15.80\% & 32.58\% & 16.38\% & 30.75\% & 13.59\% & 27.00\%\\
\midrule
Norm + Spontaneous Speech + Spontaneous KenLM & 6.05\% & 15.52\% & 16.58\% & 36.98\% & 16.91\% & 33.45\% & 14.11\% & 29.85\%\\
\midrule
Norm + Prepared Speech + Early Stopping & \textbf{4.54\%} & \textbf{12.96\%} & 15.73\% & 35.97\% & 14.75\% & 30.16\% & \textbf{12.45\%} & \textbf{27.31\%}\\
\midrule  
Norm + Spontaneous Speech + Early Stopping & 4.59\% & 13.14\% & 15.97\% & 36.52\% & \textbf{14.66\%} & \textbf{29.83\%} & 12.47\% & 27.33\%\\
\bottomrule  
\end{tabular} 
\caption{Additional Experiments. All the experiments used the Wav2vec 2.0 XLSR-53 pre-trained model for finetuning.}
\label{tab:adddevset}
\end{table*}

\section{Conclusions}\label{sec:conclusions}

In this work we presented our efforts to build a robust ASR model using multiple approaches, such as selective noise insertion and domain specific fine-tuning. In our experiments we found that fine-tuning a strong baseline with additional public available data in multiple domains and using normalization, even for a few epochs, can improve performance. With our results we were able to improve on the test set in 3 of the 4 tracks available over the strong baseline provided.

As future works, we plan to train a ASR model using a dynamic noise insertion approach that do not depend on choosing specific datasets previously.

\section*{Acknowledgments}

This research was funded by CEIA with support by the Goi\'{a}s State Foundation (FAPEG grant \#201910267000527)\footnote{\url{http://centrodeia.org/}}. We also would like to thank Cyberlabs Group\footnote{\url{https://cyberlabs.ai/}} for the support for this work. 

\bibliographystyle{unsrt}  
\bibliography{references}  

\end{document}